\documentclass[manuscript,screen,nonacm]{acmart}
\AtBeginDocument{%
  \providecommand\BibTeX{{%
    \normalfont B\kern-0.5em{\scshape i\kern-0.25em b}\kern-0.8em\TeX}}}

\copyrightyear{2022} 
\acmYear{2022} 
\setcopyright{licensedothergov}\acmConference[FAccT '22]{2022 ACM Conference on Fairness, Accountability, and Transparency}{June 21--24, 2022}{Seoul, Republic of Korea}
\acmBooktitle{2022 ACM Conference on Fairness, Accountability, and Transparency (FAccT '22), June 21--24, 2022, Seoul, Republic of Korea}
\acmPrice{15.00}
\acmDOI{10.1145/3531146.3533192}
\acmISBN{978-1-4503-9352-2/22/06}

\usepackage{microtype}
\usepackage{xcolor}
\usepackage{multirow}
\usepackage{color}

\usepackage{subcaption}
\DeclareMathOperator*{\argmax}{arg\,max}

\begin{document}


\title[Can Machines Help Us Answering Question 16 in Datasheets?]{Can Machines Help Us Answering Question 16 in Datasheets, and In Turn Reflecting on Inappropriate Content?}

\author{Patrick Schramowski}
\email{schramowski@cs.tu-darmstadt.de}
\affiliation{%
  \institution{%Computer Science Department, 
  Technical University Darmstadt, %and
  %TU Darmstadt,
Hessian Center for AI %(hessian.AI)
}
  %\streetaddress{P.O. Box 1212}
  \city{Darmstadt}
  \country{Germany}
}
\author{Christopher Tauchmann}
\email{tauchmann@cs.tu-darmstadt.de}
\affiliation{%
  \institution{%Computer Science Department, 
  Technical University Darmstadt, Hessian Center for AI %(hessian.AI)
  }
  %\streetaddress{P.O. Box 1212}
  \city{Darmstadt}
  \country{Germany}
}
\author{Kristian Kersting}
%\authornotemark[1]00,0,
\email{kersting@cs.tu-darmstadt.de}
\affiliation{%
  \institution{%Computer Science Department and Centre for Cognitive,
  %TU Darmstadt,
Technical University Darmstadt,
Centre for Cognitive Science Darmstadt, %and 
Hessian Center for AI %(hessian.AI)
}
  %\streetaddress{P.O. Box 1212}
  \city{Darmstadt}
  \country{Germany}
}

\renewcommand{\shortauthors}{Schramowski, et al.}

\begin{abstract}
 {\textcolor{purple}{
 {\fontencoding{U}\fontfamily{futs}\selectfont\char 49\relax}
 This paper contains images and descriptions that are offensive in nature.
 }}
  
Large datasets underlying much of current machine learning raise serious issues concerning inappropriate content such as offensive, insulting, threatening, or might otherwise cause anxiety. 
This calls for increased dataset documentation, e.g., using datasheets. They, among other topics, encourage to reflect on the composition of the datasets. So far, this documentation, however, is done manually and therefore can be tedious and error-prone, especially for large image datasets.
Here we ask the arguably ``circular'' question of whether a machine can help us reflect on inappropriate content, answering Question 16 in Datasheets.
To this end, we propose to use the information stored in pre-trained transformer models to assist us in the documentation process.
Specifically, prompt-tuning based on a dataset of socio-moral values steers CLIP to identify potentially inappropriate content, therefore reducing human labor. We then document the inappropriate images found using word clouds, based on captions generated using a vision-language model.
The documentations of two popular, large-scale computer vision datasets---ImageNet and OpenImages---produced this way suggest that machines can indeed help dataset creators to answer Question 16 on inappropriate image content.

\end{abstract}

%\CopyrightYear{2022} 
%\conferenceinfo{FAccT '22,}{June 21--24, 2022, Seoul, Republic of Korea}
%\isbn{978-1-4503-9352-2/22/06}\acmPrice{$15.00}
%\doi{https://doi.org/10.1145/3531146.3533192}



\maketitle

\section{Introduction}
Transfer learning from models that have been pre-trained on huge datasets has become standard practice in many computer vision (CV) and natural language processing (NLP) tasks and applications. While approaches like semi-supervised sequence learning \cite{dai2015semi} and datasets such as ImageNet \cite{deng2009imagenet}---especially the ImageNet-ILSVRC-2012 dataset with 1.2 million images--- established pre-training approaches, the training data size increased rapidly to billions of training examples \cite{brown2020Language, jia2021align}, steadily improving the capabilities of deep models.
Recent transformer architectures with different objectives such as autoregressive \cite{radford2019language} and masked  \cite{devlin2018bert} language modeling as well as natural language guided vision models \cite{radford21Learning} for multi-modal vision-language (VL) modeling have even enabled zero-shot transfer to downstream tasks, avoiding the need for 
task-specific fine-tuning.

However, in all areas, the training data in the form of large and undercurated, internet-based datasets is problematic involving, e.g., stereotypical and derogatory associations \cite{gebru21datasheets, bender2021Stochastic}.
Along this line, \citet{gebru21datasheets} described dominant and hegemonic views, which further harm marginalized populations, urging researchers and dataset creators to invest significant resources towards dataset curation and documentation. Consequently, the creation of datasheets became common practice when novel datasets such as \cite{desai2021redcaps} were introduced. However, the documentation of \citet{desai2021redcaps} shows that careful manual documentation is difficult, if not even unfeasible, due to the immense size of current datasets: `\textit{We manually checked 50K} [out of 12M] \textit{random images in RedCaps and found one image containing
nudity (exposed buttocks; no identifiable face)}'. Also, in the process of creating a datasheet for the BookCorpus, \citet{bandy21addressing} stated that further research is necessary to explore the detection of potential inappropriate concepts in text data.
\citet{birhane2021Large} manually checked for and found misogynistic and pornographic in several common CV datasets.
However, misogynistic images and pornographic content are only part of the broader concept of inappropriate content. It remains challenging to identify concepts such as general offensiveness in images, including abusive, indecent, obscene, or menacing content.

To make a step towards meeting the challenge, the present work proposes a semi-automatic method, called Q16, to document inappropriate image content. We use the VL model CLIP \cite{radford21Learning} to show that it is indeed possible to (1) steer pre-trained models towards identifying inappropriate content as well as (2) the pre-trained models themselves towards mitigating the associated risks. In the Q16 setup, prompt-tuning steers CLIP to detect inappropriateness in images. Additionally, Q16 employs the recent autoregressive caption generation model MAGMA \cite{Eichenberg_Black_Weinbach_Parcalabescu_Frank_2021} to provide accessible documentation.
Thus, Q16 assists dataset documentation and curation by answering Question 16 of \cite{gebru21datasheets}, which also explains its name: \textit{Does the dataset contain data that, if viewed directly, might be offensive, insulting, threatening, or might otherwise cause anxiety?}

\begin{figure*}[t]
    \centering
    \includegraphics[width=.85\linewidth]{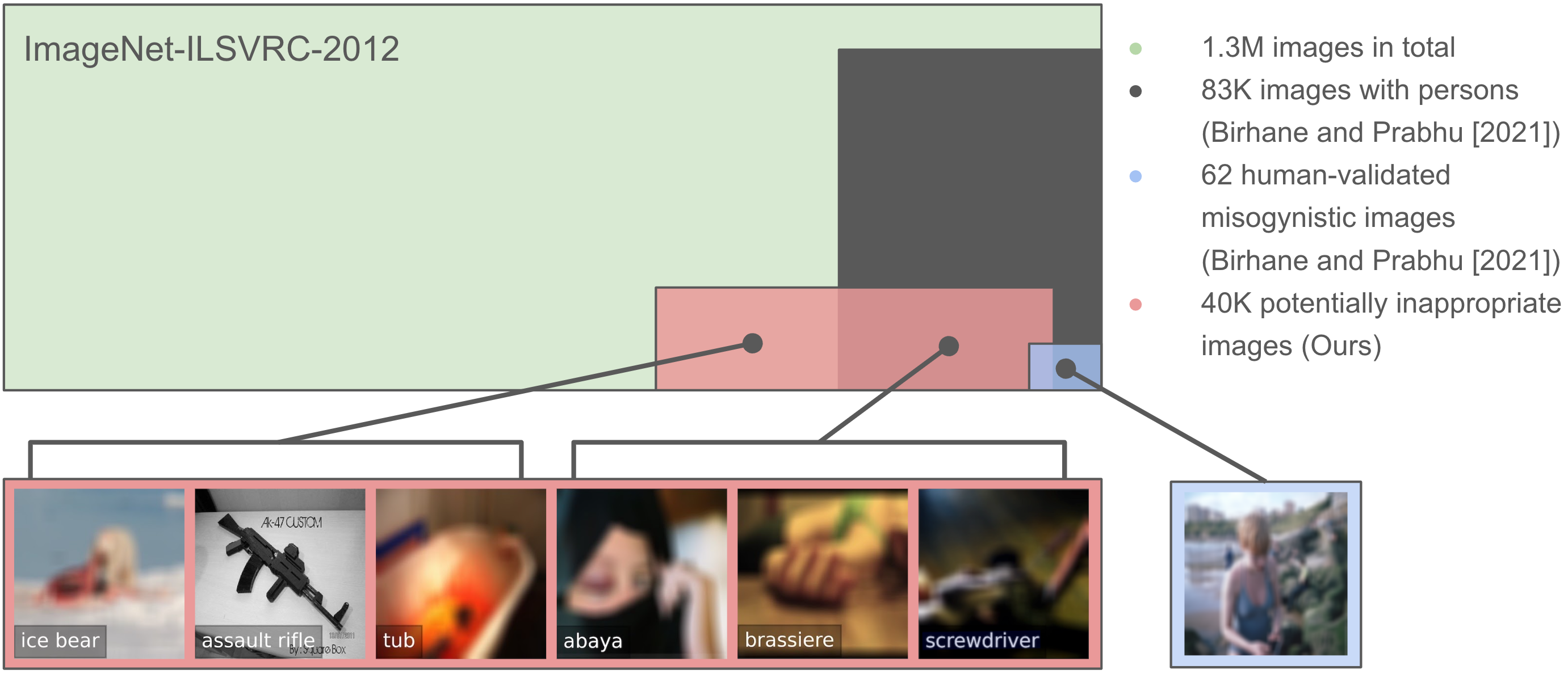}
    \caption{Range of identified inappropriate concepts illustrated using ImageNet (green). The other colors refer to different data-subsets: a selection of all images displaying persons (dark gray), potentially inappropriate images identified by our approach (red), and human-validated inappropriate (misogynistic) images identified in the study of \citet{birhane2021Large} (blue). The detected images in our approach partly overlap with the one in blue.
    Sizes are only illustrative, and actual numbers are given in the legend (right). Due to their apparent offensive content, we blurred the images.}
    \Description{Range of identified inappropriate concepts using our Q16 approach illustrated using ImageNet. Beside scenes with persons these concepts also include objects, text and symbols.}
    \label{fig:inappropriate_concepts}
\end{figure*}

We illustrate Q16 on the popular ImageNet-ILSVRC-2012 \cite{deng2009imagenet} and OpenImages \cite{Kuznetsova2020theopen} dataset and show that large computer vision datasets contain additional inappropriate content, which previous documentations, such as \cite{birhane2021Large}, had not detected, cf.~Fig.~\ref{fig:inappropriate_concepts}. In contrast to images identified in previous approaches, e.g., images showing nudity and misogynistic images (blue), Q16 detects a larger and broader range of potential inappropriate images (red). These images show violence, misogyny, and otherwise offensive material. Importantly, this includes images portraying persons (dark gray) as well as objects, symbols, and text. 

The rest of the paper is organized as follows. We start off with a brief overview of related work and required background introducing pre-trained models and their successes as well as concerns raised. 
Next, we describe inappropriate image content and show that common deep models cannot reliably detect potential inappropriate images due to the lack of sufficient data. We then continue by demonstrating that recent models, guided by natural language during the pre-training phase, can classify and describe inappropriate material based on their retained knowledge.
Before concluding, we present our automated dataset documentation exemplary on the ImageNet-ILSVRC-2012 and OpenImagesV6 datasets.
We provide our models and the necessary data and code to reproduce our experiments and utilize our proposed method.\footnote{\url{https://github.com/ml-research/Q16}}

\section{Background and related work}
\label{sec:background}
In this section, we describe pre-trained models in NLP, CV, and recent VL models.
Furthermore, we touch upon related work aiming to improve dataset documentation and curation as well as identifying problematic content in datasets. 
%BACKGROUND
\subsection{Large pre-trained models} \label{subsec:ptm}
Large-scale transformer-based language models revolutionized many NLP tasks \citep{Lin_Wang_Liu_Qiu_2021}.
As large, pre-trained models form the backbone of both natural language processing and computer vision today, it is natural that multimodal vision-language models \cite{radford21Learning, ramesh2021zero, jia2021align} extend these lines of research.

For their CLIP model, \citet{radford21Learning} collected over 400M image-text pairs (WebImageText dataset) to show that the success in large-scale transformer models in NLP can be transferred to vision and multimodal settings. One major takeaway from their work is the benefit of jointly training an image encoder and a text encoder to predict the correct pairings of a batch of (image, text) training examples. Typical vision models \cite{he2016deep, tan19EfficientNet} jointly train an image feature extractor and a classifier. \citet{radford21Learning}, the authors of CLIP, proposed to synthesize the learned text encoder with a (zero-shot) linear classifier at test time by embedding the names or descriptions of the target dataset's classes, e.g. ``The image shows \textit{$<$label$>$}.'', thus reducing the (computational) cost of fine-tuning the model and using it as it was trained.
Such models and their zero-shot capabilities display significant promise for widely-applicable tasks like image retrieval or search. The relative ease of steering CLIP toward various applications with little or no additional data or training unlocks novel applications that were difficult to solve with previous methods, e.g., as we show, classify potential inappropriate image content.
\subsection{Issues arising from large datasets}
Large-scale models require a tremendous amount of training data. The most recent and successful models, such as GPT-3 \cite{brown2020Language}, CLIP \cite{radford21Learning}, DALL-E \cite{ramesh2021zero} and other similar models, are trained on data scraped from the web, e.g. using CommonCrawl. The information they acquire from this data is largely uncontrolled. However, even ImageNet \cite{deng2009imagenet}, which was released in 2012 and remains one of the most popular datasets in the computer vision domain to this day \cite{brock21high, tan21efficentnetv2}, contains questionable content \cite{birhane2021Large}.
The entailed issues have been discussed for language models, for instance, models producing stereotypical and derogatory content \cite{bender2021Stochastic}, and for vision model respectively CV datasets highlighting, e.g., gender and racial biases \cite{steed21image, Larrazabal12592, wang20revise, denton21on}.

Consequently, \citet{gebru21datasheets} urged the creation of datasheets accompanying the introduction of novel datasets including a variety of information on the dataset to increase transparency and accountability within the ML community, and most importantly, help researchers and practitioners to select more appropriate datasets for their tasks.
The documentation and curation of datasets have become a very active research area, and along with it, the detection of inappropriate material contained in datasets and reflected by deep models.

\citet{dodge_documenting_2021} documented the very large C4 corpus with features such as `text source' and `content', arguing for different levels of documentation. They also address how C4 was created and show that this process removed texts from and about minorities. 
A vast body of work to date that describes methodologies to tackle, abusive, offensive, hateful \cite{Glavas_Karan_Vulic_2020}, toxic \cite{Han_Tsvetkov_2020}, stereotypical \cite{Nadeem_Bethke_Reddy_2021} or otherwise biased content \cite{Dhamala_Sun_Kumar_Krishna_Pruksachatkun_Chang_Gupta_2021} come from NLP. For several years, workshops on language\footnote{https://aclanthology.org/volumes/W17-30/} and offensive\footnote{https://sites.google.com/site/offensevalsharedtask/home} language are carried out, producing evaluation datasets. Furthermore, Google hosts an API for the automatic detection of toxicity\footnote{https://www.perspectiveapi.com/} in language, and research introduced toxicity benchmarks for generative text models \cite{gehman2020realtoxicityprompts}.
Additionally, the definitions and datasets on such tasks as bias- and hate-speech identification become increasingly complex \cite{Sap_Gabriel_Qin_Jurafsky_Smith_Choi_2020}.
Accordingly, most of the research on automatic methods focuses solely on text. 

With the present study, we aim to push the development of methods for the CV domain. 
\citet{yang_towards_2020} argued towards fairer datasets and filter parts of ImageNet. Specifically, they see issues in ImageNet's concept vocabulary based on WordNet and include images for all concept categories (some hard to visualize). Furthermore, the inequality of representation (such as gender and race) in the images that illustrate these concepts is problematic.
\citet{birhane2021Large} provided modules to detect faces and post-process them to provide privacy, as well as a pornographic content classifier to remove inappropriate images. 
Furthermore, they conducted a hand-surveyed image selection to identify misogynistic images in the ImageNet-ILSVRC-2012 (ImageNet1k) dataset.
\citet{gandhi2020scalable} aimed to detect offensive product content using machine learning; however, they have described the lack of adequate training data.
Recently, \citet{nichol21glide} applied CLIP to filter images of violent objects but also images portraying people and faces.

\subsection{Retained knowledge of large models}
Besides the performance gains, large-scale models show surprisingly strong abilities to recall factual knowledge from the training data \cite{petroni2019language}. For example, \citet{roberts2020how} showed large-scale pre-trained language models' capability to store and retrieve knowledge scales with model size. 
\citet{schick2021self} demonstrated that language models can self-debias the text they produce, specifically regarding toxic output. 
Similar to our work, they prompt a model. However, they use templates with questions in the form of ``this model contains <MASK>'', where the gap is filled with attributes, such as toxicity, whereas we automatically learn prompts.
Furthermore, \citet{MCM} and \citet{schramowski2020themoral} showed that the retained knowledge of such models carries information about moral norms aligning with the human sense of \textit{``right''} and \textit{``wrong''} expressed in language. 
Similar to \cite{schick2021self}, \citet{schramowski2022large} demonstrated how to utilize this knowledge to guide autoregressive language models' text generation to prevent their toxic degeneration.

%OUR METHOD
\section{The Q16 Pipeline for Datasheets}
\begin{figure*}[t]
    \centering
    \includegraphics[width=0.84\textwidth]{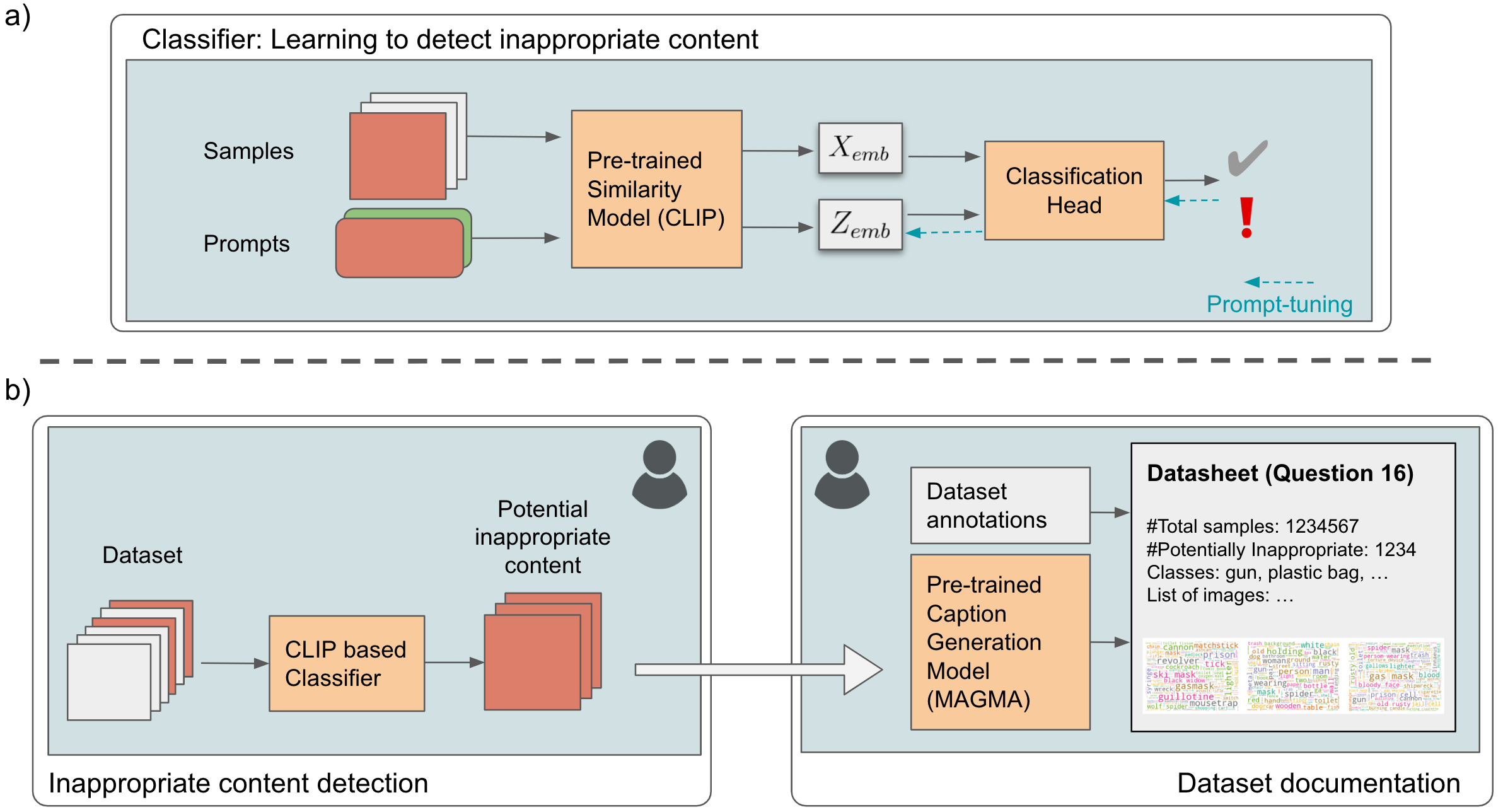}
    \caption{Overview of the Q16 pipeline, a two-step dataset documentation approach. a) In order to utilize the implicit knowledge of the large pre-trained models, prompt-tuning steers CLIP to classify inappropriate image content. b) Dataset documentation process: First, a subset with potentially inappropriate content is identified. Secondly, these images are documented by, if available, image annotations and automatically generated image descriptions. Both steps are designed for human interaction. 
    }
    \Description{Overview of the Q16 pipeline, a two-step dataset documentation approach. First, a subset with potentially inappropriate content is identified. Secondly, these images are documented.}
    \label{fig:method}
\end{figure*}
Let us now start to introduce our semi-automatic method to document inappropriate image content. To this end, we first clarify the term ``inappropriateness'' in Sec.~\ref{sec:inappr_content_definition}. 
Then we present and evaluate models, including our approach (illustrated in Fig~\ref{fig:method}a), to classify inappropriate image content. 
Specifically, Fig~\ref{fig:method}a shows a dataset representing socio-moral norms, which will be detailed in Sec.~\ref{subsec:smid}, steering CLIP to detect inappropriate content using (soft)~prompt-tuning (cf.~Sec.~\ref{sec:inappr_classifier}).

Lastly, in Sec.~\ref{sec:documentation}, we present the two-step semi-automated documentation (cf.~Fig~\ref{fig:method}b).
Notably, both steps include human interaction. 
First, CLIP and the learned prompts from Fig~\ref{fig:method}a are used to detect inappropriate images within the dataset.
Detection is conservative, aiming to identify all potentially inappropriate content. Accordingly, the subsets are of considerable size, e.g., ~40K in the case of ImageNet1k. Therefore, the second step generates automatic image descriptions to assist the dataset creators in describing and validating the identified content.
The final documentation of Q16 includes the ratio of identified images and the total amount of samples, and a summary of the identified concepts.
To overview the contained concepts in an easily accessible way, we generate word clouds based on two properties: the dataset annotation and generated description.

\subsection{Inappropriate image content.}
\label{sec:inappr_content_definition}
Let us start off by clarifying the way we use the term ``inappropriate'' in our work and describing the term in the context of images. Question 16 of 
Datasheets for Datasets \cite{gebru21datasheets} asks one to document the dataset composition regarding the contained ``data that, if viewed directly, might be offensive, insulting, threatening, or might otherwise cause anxiety''. 
Consequently, \citet{birhane2021Large} applied different models to detect visible faces (thus violating privacy rights) and pornographic content. Additionally, they conducted a survey identifying misogynistic images. However, the definition of \citet{gebru21datasheets} includes a broader range of inappropriate concepts not addressed by current work.

According to the Cambridge dictionary\footnote{https://dictionary.cambridge.org/dictionary/english/offending%, accessed on 3rd October 2021
}, `offending' can be phrased as `unwanted, often because unpleasant and causing problems'. Additionally, in the context of images and text, according to Law Insider\footnote{https://www.lawinsider.com/dictionary/offending-materials%, accessed on 3rd October 2021
}:
`\textit{Offending Materials means any material, data, images, or information which is (a) in breach of any law, regulation, code of practice or acceptable use policy; or (b) defamatory, false, inaccurate, abusive, indecent, obscene or menacing or otherwise offensive; or (c) in breach of confidence, copyright or other intellectual property rights, privacy or any other right of any third party.}' In the present paper, we focus on images following the definition (b). This definition aligns with definitions of previous work detecting hate speech \cite{gomez2020exploring} and offensive product images \cite{gandhi2020scalable}.

Note that inappropriateness, especially offensiveness, is a concept that is based on social norms, and people have diverse sentiments. In the present study, we detect inappropriate content based on the implicit knowledge contained in CLIP steered with selected data (described in the following section). Therefore, the investigated `inappropriateness' may primarily surface from the group of people that have generated the selected data and the annotators but also the pre-trained model's retained knowledge. 

%dataset
\subsection{The Socio-Moral Image Database (SMID)} \label{subsec:smid}
Besides utilizing the `knowledge' of pre-trained models on inappropriate concepts, we further steer the model towards detecting (morally) inappropriate image concepts indirectly via training examples. 
I.e. we aim to find a compass guiding the encoded knowledge of CLIP and by that be able to classify inappropriate content beyond the examples shown in the steering dataset.
To this end, we propose to use the Socio-Moral Image Database (SMID) \cite{crone2018TheSocio} together with the few-shot capabilities of CLIP.
This dataset will not only be used to steer CLIP but also to evaluate the the classifier's performance in the following sections.

The SMID dataset contains $2{,}941$ images covering both morally positive and negative poles (962 negative images and 712 positive images) over several content dimensions, including objects, symbols as well as actions. 
Stimuli span the entire moral spectrum ranging from positive to negative, see Appendix Sec.~A %\ref{appendix_sec:smiddistribution}
for more details. In total, over 50 concepts are included, with negative ones such as \textit{Harm, Inequality, Degradation, Discrimination, and Exploitation}. The complete list is provided in Table~2 of \cite{crone2018TheSocio}.

The images were collected in a multi-step process and annotated by $2{,}716$ annotators. 
\citet{crone2018TheSocio} suggested to divide the data into \textit{good} (mean rating $>\!3.5$), \textit{bad} (mean rating $<\!2.5$), and neutral (rest) images.
According to this division we considered a $\text{rating}\!<\!2.5$ as (morally) inappropriate, and $\text{rating}\!>\!3.5$ as counterexamples.  %shows the density distribution of the annotated data.

% Notes from the Crone paper
% - e five moral foundations, and (4) provide stimuli that span the entire moral spectrum, from negative to positive,
% 476 participants for image collection task, half female, 10% of participants were recruited from India, with the rest from the United States.
% and additional images from researchers
% f 285 AMT workers screened resulting images
% Concepts were both positively and negatively valenced, and spanned a wide range of moral content. In total 50 concepts, such as Care / Harm, Inequality, Degradation, Discrimination, Exploitation, Equality, Politeness, Fairness, Generosity. A full list of concepts is provided in Table 2 of .

% Fig 3 was placed here (fig:smid_data_pca)
%
%

\subsection{Inappropriate content detection of Q16}
\label{sec:inappr_classifier}
\begin{table*}[t]
    \centering
    %\small
    %{\def\arraystretch{1.12}\tabcolsep=12.5pt
    \begin{tabular}{c|l||c|c|c|c}
          Architecture &Pre-training dataset &   Accuracy (\%) & Precision & Recall & F1-Score  \\ \hline
          \multirow{5}{*}{ResNet50}&\multirow{2}{*}{ImageNet1k} &$78.36\pm1.76$&$0.75\pm0.05$& $0.74\pm0.09$& $0.76\pm0.02$\\
          &&$80.81\pm2.95$&$0.75\pm0.02$& $0.81\pm0.02$& $0.80\pm0.03$\\\cline{2-6}
          &\multirow{2}{*}{ImageNet21k} &$82.11\pm1.94$&$0.78\pm0.02$& $0.80\pm0.05$& $0.78\pm0.04$\\
          & &$84.99\pm1.95$&$0.82\pm0.01$& $0.85\pm0.06$& $0.82\pm0.04$\\\cline{2-6}
          &WebImageText &$\circ90.57 \pm 1.82$&$\circ0.91\pm0.03$& $\circ0.89\pm0.01$& $\circ0.88\pm0.03$ \\ \hline
          ViT-B/32&WebImageText &$94.52 \pm 2.10$&$0.94\pm0.04$& $0.91\pm0.02$& $0.92\pm0.01$ \\
          ViT-B/16&WebImageText &$\mathbf{\bullet96.30 \pm 1.09}$&$\mathbf{\bullet0.95\pm0.02}$& $\mathbf{\bullet0.97\pm0.01}$& $\mathbf{\bullet0.97\pm0.02}$
    \end{tabular}
    %}
    \caption{Performances of pre-trained models ResNet50 and ViT-B. CLIP-based models trained on WebImageText outperform baselines and show remarkable performance in identifying the inappropriate content contained in SMID. The ResNet50 is pre-trained on ImageNet1k, ImageNet21k \cite{deng2009imagenet} and the WebTextImage dataset \cite{radford21Learning}. The ViT is pre-trained on the WebTextImage dataset. On the ImageNet datasets, we applied linear probing (top) and fine-tuning (bottom), and on the WebImageText-based models, soft-prompt tuning. The overall best results are highlighted \textbf{bold} with the $\bullet$ marker and best on the ResNet50 architecture with $\circ$ markers. Mean values and standard deviations are reported.}
    \label{tab:smid_classification_results}
\end{table*}
Let us now move on to presenting and evaluating different models, including our  CLIP-based Q16 approach, for the task at hand to classify inappropriate image content. Here inappropriate content is defined by the SMID data and annotation, see section above.
In the following experiments, $10$-fold cross-validated results are reported.
%Additional to the following quantitative results we provide an analyze of the pre-trained models' embedding spaces in the Appendix Sec.~\ref{appendix_sec:pca}

\subsubsection{Deep Learning baselines.} 
As baselines we fine-tuned two standard pre-trained CV models (PMs) to investigate how well deep neural networks can identify inappropriate content.
Similar to \citet{gandhi2020scalable}, we used the ResNet50 architecture \cite{he2016deep}, pre-trained on ImageNet datasets \cite{deng2009imagenet}.

Tab.~\ref{tab:smid_classification_results} shows the performance of both the fine-tuned model (training all model parameters) and a model with only one linear probing layer. In our work, the probing layer refers to adding one final classification layer to the model.
This part of the table shows inconclusive results: even if the performance increases when a larger dataset (ImageNet21k) is used. After fine-tuning the whole model, recall increases; precision, however, is still comparatively low. Specifically, the resulting low precision and low recall of the linear probed ImagNet1k-based models show problems classifying truly inappropriate images as well as distinguishing between truly non-inappropriate and inappropriate images.
We will use these models as baselines to investigate if more advanced PMs (trained on larger unfiltered datasets) carry information about potential inappropriate image content.

\subsubsection{Zero and few-shot capabilities of CLIP to infer inappropriate content.}
Next, we will investigate if CLIP's contrastive pre-training step contains image-text pairs that equip the model with a notion of inappropriate concepts. 

Due to the natural language supervision, CLIP should implicitly have acquired knowledge about what a human could ---depending on the context--- perceive as inappropriate content.
We confirmed this assumption %CLIP's Vision-Transformer (ViT) can indeed distinguish inappropriate content and corresponding counterexamples without being explicitly trained to do so, encoding task-specific knowledge 
with a PCA of the embedded images (see Appendix Sec.~B %\ref{appendix_sec:pca} 
for details)
%Note that based on a PCA of the embedded images (see Appendix Sec.~\ref{appendix_sec:pca} for details), we can already see that CLIP's Vision-Transformer (ViT) can indeed distinguish inappropriate content and corresponding counterexamples without being explicitly trained to do so, encoding task-specific knowledge.
%This observation confirms our assumption that due to the natural language supervision, CLIP implicitly acquired knowledge about what a human could ---depending on the context--- perceive as inappropriate content.

Now, the inappropriateness classifier of our approach (Fig.~\ref{fig:method}a) utilizes this `knowledge'. It is based on prompting CLIP with a natural language sentence. Our prompts have the form ``This image is about something $<$\textit{label}$>$.'', helping to specify that the text is actually about the content of the image.
To map the labels of the SMID dataset to natural language sentences, we used the following labels according to \citet{crone2018TheSocio}:
\textit{bad/good behavior}, \textit{blameworthy/praiseworthy}, 
\textit{positive/negative} and \textit{moral/immoral}. The \textit{positive} and \textit{negative} labels resulted in the best zero-shot performance.
Images are encoded via the pre-trained visual encoder, similar to the ResNet50 model. However, instead of training a linear classifier to obtain class predictions as in these models, we now operate on the similarity of samples (the cosine similarity) in the representation space:
\begin{equation}
    Sim(x, z) = \frac{E_{visual}(x) * E_{text}(z)}{||E_{visual}(x)||_2 * ||E_{text}(z)||_2} \ ,
\end{equation}
where $E_{visual}$ and $E_{text}$ are the visual and text encoders, $x$ is an image sample and $z$ a prompt. 
\begin{figure*}[t]
    \centering
    \includegraphics[width=0.9\linewidth]{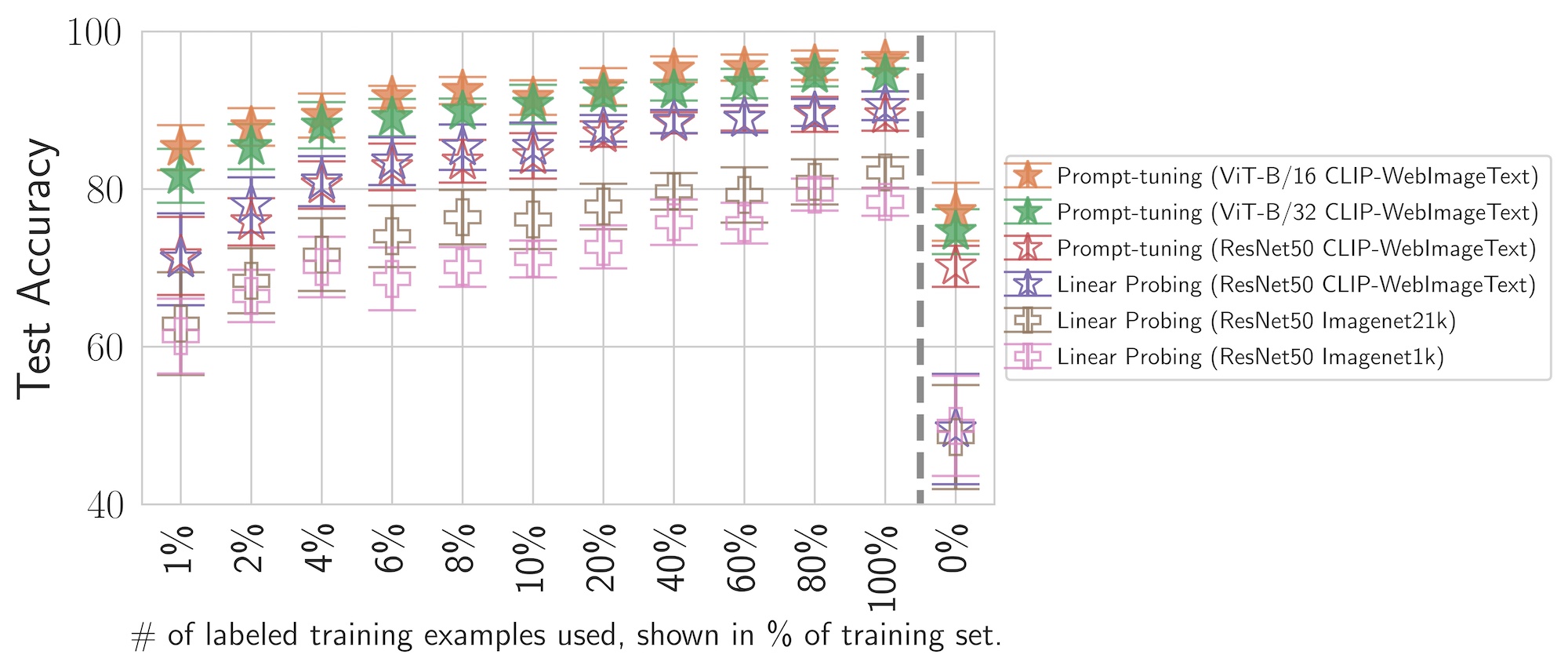}
    \caption{Performance of pre-trained models ResNet50 and ViT-B. 
    CLIP-based models outperform baselines and show remarkable (zero- and few-shot) performances in identifying the inappropriate content contained in SMID.
    ResNet50 is pre-trained on ImageNet1k, ImageNet21k \cite{deng2009imagenet} and the WebTextImage dataset \cite{radford21Learning}. ViT is pre-trained on the WebTextImage dataset. On the ImageNet datasets, we applied linear probing (top), and on the WebImageText-based models used soft-prompt tuning. Tuning was performed on different sizes of the SMID training set where $100\%$ corresponds to $1{,}506$ images. 
    }
    \Description{Performance of pre-trained models ResNet50 and ViT-B. The models are evaluated on the SMID dataset. CLIP-based models outperform baselines and show remarkable (zero- and few-shot) performances in identifying the inappropriate content contained in SMID.}
    \label{fig:smid_cv_results}
\end{figure*}
Fig.~\ref{fig:smid_cv_results} ($0\%$, prompt-tuning) shows that this approach already performs on par with the ImageNet-based PMs fine-tuned on SMID ($100\%$, linear probing). 
However, the zero-shot approach can classify true-negative samples well but performs not so well on classifying positives. This observation suggests that both prompts, at least the one corresponding to the positive class label, are not optimal.

\subsubsection{Steering CLIP to infer inappropriate content via prompt-tuning.} \label{par:steering}
The manual handwritten prompts may not be the best way to query the model. Consequently, we used prompt-tuning \citep{lester_power_2021, qin2021learning, Hambardzumyan_Khachatrian_May_2021} to learn continuous optimal prompts. Prompt-tuning optimizes the prompts by searching for the optimal text embeddings for a given class label.

Several variations employ prompt-tuning: Prefix-tuning, for example, learns a prefix to add to a sample's embedding \citep{qin2021learning} on every model layer. \citet{lester_power_2021} created new (prompt) embeddings only once by pre-pending a small vector to the original input embedding for all downstream examples. \citet{Hambardzumyan_Khachatrian_May_2021} updated both the input and final embeddings once.
In contrast, we propose to learn the entire final sentence embedding once, obtaining one sentence embedding, $z_{emb}$, for each class label. In turn, the distinction of inappropriate and other images is defined as an optimization task using gradient descent as follows:
\begin{equation}
\mathbf{\hat{z}}_{emb} = \argmax\nolimits_{\mathbf{z}_{emb}}\{L(\mathbf{z}_{emb})\} \ ,
\end{equation} where
\begin{align}
    L(\mathbf{z}_{emb}) &= - \frac{1}{|X|}\sum\nolimits_{\mathbf{x} \in X} \mathbf{y} \ \log(\mathbf{\hat{y}}) \text{, } \\ \text{with } \mathbf{\hat{y}} &= \text{softmax}(Sim(\mathbf{x},\mathbf{z}_{emb})) \text{ .}
\end{align}
Here, the parameters $\theta$ of $E_{visual}$ and $E_{text}$ are not updated. The initial prompts $Z$ are only propagated through $E_{text}$ once and the resulting embeddings $\mathbf{z}_{emb} \in Z_{emb}$ are optimized. Furthermore, $\mathbf{y}$ is the class label, and $X$ a batch in the stochastic gradient descent optimization. 
% Fig 5 was placed here: fig:prompttuning
Our prompt-tuning approach is summarized visually in Fig.~\ref{fig:method}; further details on applying it to the SMID dataset can be found in Appendix Sec.~C.%\ref{appendix_sec:prompttuning}.
% Moved to appendix
%Moreover, Fig.~\ref{fig:prompttuning} shows exemplary nearest image neighbors of the learned prompts. The image on the right side clearly portrays possible inappropriate content. In contrast, the image on the left side displays a positive scene as a counterexample. 
%

Fig.~\ref{fig:smid_cv_results} also shows an evaluation of CLIP using the soft prompts (prompt-tuning). We can see that a small portion of the training data (e.g., 4\%, 60 images) already leads to an increase of the vision transformer's (ViT-B) performance to over 90\%. 
This shows that indeed large pre-trained model can be steered more efficient, i.e. generalize and detect inappropriate concepts beyond the training samples.
In general, the ViT-B outperforms the pre-trained ResNet50 models. Furthermore, ViT-B/16 outperforms the ViT-B/32, indicating that not only the dataset's size is important, but also the capacity of the model (ViT-B/16 has higher hidden-state resolution than
the ViT-B/32). 
Training ViT with the full training set results in $96.30\% \pm 1.09$ (cf. Tab.~\ref{tab:smid_classification_results}) accuracy. 

Overall, one can see that steering CLIP towards inferring potentially inappropriate concepts in images requires only little additional data. In contrast to other pre-trained models, it provides a reliable method to detect inappropriate images.

\subsection{Dataset documentation of Q16}
\label{sec:documentation}
Using our prompt-tuning approach, the pre-selection by CLIP can, in principle, extract possible inappropriate images automatically that can then be used for dataset documentation. 
However, we have to be %a bit more 
careful since inappropriateness is subjective to the user---e.g., researchers and practitioners selecting the dataset for their tasks---and, importantly, to the task at hand. In our case, the steered model may primarily mirror the moral compass and social expectations of the $2{,}716$ annotators.
Therefore, it is required that humans and machines interact with each other, and the user can select the images based on their given setting and requirements. 
Hence, we do not advise removing specific images but investigating the range of examples and inappropriate content selected by the model and thereby documenting the dataset.
In the following, we present our approach to assist data creators not only in identifying but also describing the identified potential inappropriate content.

% new based on reviews
Fig.~\ref{fig:method}b shows our human-in-the-loop, (\textit{cf.} Sec.~\ref{subsec:answeringq16}), documentation setting. 
The above mentioned detection is conservative, aiming to identify all potentially inappropriate content. Accordingly, the subsets are of considerable size, e.g., ~40K in the case of ImageNet1k. Therefore, the first step to assist the dataset creators in describing and validating the identified content is the automatic generation of image descriptions, \textit{cf.} Sec.~\ref{subsec:captiongeneration}.
To overview the contained concepts in an easily accessible way, we generate word clouds based on two properties: the dataset annotation and generated description, \textit{cf.} Sec.~\ref{subsec:wordcloudgeneration}.

\subsubsection{Answering Datasheet Question 16: Does the dataset contain data that, if viewed directly, might be offensive, insulting, threatening, or
might otherwise cause anxiety?}
\label{subsec:answeringq16}
As intended by the original datasheets paper \cite{gebru21datasheets}, dataset creators should start describing the curation process concerning this question.
Whereas our approach could also be used for the curation, we focus solely on documenting the final dataset content to mitigate unwanted societal biases in ML models, and help users select appropriate datasets for their chosen tasks.

The dataset documentation should contain the total amount of images and the ratio of identified, potentially inappropriate images. Since the process of creating a datasheet is not intended to be automated \cite{gebru21datasheets}---however, the quality of current datasheets \cite{desai2021redcaps} indicate that a semi-automated method is unavoidable---, the resulting subset should be manually validated and described by the dataset's creators. Our approach aims to reduce impractical human labor while encouraging creators to reflect on the process carefully. 

\subsubsection{Automatic caption generation.}
\label{subsec:captiongeneration}
In order to categorize and thus describe the identified content, dataset annotations can be used if they are available.
However, these annotations often may not describe the complete image content, especially in the case of natural images. Therefore, we utilize automatic generation of image descriptions, cf. Fig.~\ref{fig:method}b (right).
To this end, we propose to generate text using a caption-generation model. Specifically, we used MAGMA (Multimodal Augmentation of Generative Models) \cite{Eichenberg_Black_Weinbach_Parcalabescu_Frank_2021}. MAGMA is a recent text generation model based on multimodal few-shot learners \cite{Tsimpoukelli_Menick_Cabi}. It uses both the CLIP and GPT-J \cite{gpt-j} models and adds pre-training and fine-tuning steps on several datasets to generate image captions from image-text pairs. These captions are especially beneficial because they include the encyclopedic knowledge of GPT-J, and as such, knowledge on socio-moral norms (similar to the one we obtain from CLIP). 
Further, the multimodal input enables one to guide the resulting textual description. Since we aim to generate ``neutral'' image descriptions, we use the prompt \textit{<A picture of>} and add the output of multiple generations to the image description. To sample from the model, we applied top-k filtering. In order to acquire greater variety in the descriptions, we used different temperature values. 
\subsubsection{Word cloud generation.}
\label{subsec:wordcloudgeneration}
Actually, Question 16 asks the dataset curator to be familiar with a broad range of inappropriate concepts. Whereas our Q16 approach already helps reduce the number of inappropriate images to be checked and, in turn, human labor, even the validation of the reduced set may still require a lot of manual effort.
To provide a concise overview, we propose to compute word clouds to summarize the complex captions generated.
More precisely, we present the identified, potentially inappropriate content within the dataset using three different kinds of word clouds from dataset annotations and generated textual image descriptions. All word clouds highlight words or bi-grams based on their frequency and rank.

The first word cloud requires existing dataset annotations, e.g., class labels, and provides first insights of identified concepts and could highlight sensible labels.
The word cloud visualizes the information by highlighting the most-frequent annotations. However, note that dataset creators should also pay attention to infrequent occurrences indicating deviating concepts compared to other examples from, e.g., the same class. Many images with the same annotation could indicate a general negative association. 

Following the same procedure, the second word cloud describes the identified set of images using the generated text and thus independent of the dataset annotations. Therefore, this word cloud potentially describes identified concepts not captured by the first word cloud.

For the third word cloud, we use a chi-squared weighting of the word/bi-gram frequencies to illustrate differences between the identified inappropriate image set and the remaining images; common text descriptions occurring in both sets are removed. Each word $i$ is assigned the following weight:  
\begin{equation}
    weight_i = \frac{(observed_i - expected_i)^2}{expected_i} \ ,
\end{equation}
where $observed_i$ is the observed frequency of word $i$ in the inappropriate subset and $expected_i$ the expected value, i.e., the observed word frequency describing the dataset's remaining samples. 
This word cloud highlights the conspicuous descriptions that can be traced back to the corresponding images.

It is noteworthy that these wordclouds highlight frequent concepts for documentation purpose. Thus, it may be easy to dismiss the severity of inappropriateness if the database contains less of that particular image content, e.g. some inappropriate content may be less common but more severe. Therefore, we advice dataset creators to also inspect infrequent concepts.  

Finally, we would like to note that our pipeline also produces several statistics such as exact word frequencies and traceable image descriptions that we do not include directly in the datasheet. The dataset creators can provide this additional information as a supplement next to the identified image IDs.

\section{Answering Datasheet Question 16 for ImageNet and OpenImages}
Now we have everything together to provide an exemplary datasheet documentation, here for the CV datasets ImageNet \cite{deng2009imagenet} and OpenImages \cite{Kuznetsova2020theopen}.
To identify inappropriate content within the datasets, we used the public available ViT-B/16\footnote{In our repository we default to the even larger ViT-L/14 variant which was released after submission of this manuscript.} variant of CLIP steered by SMID-based optimized prompts. We observed that shifting the negative threshold to a rating of $1.5$ instead of $2.5$ provides a conservative but reliable classifier; hence we determined the prompts with these corresponding few-shot examples. For the documentation process we utilized the ResNet50x16 MAGMA model and generated 10 captions ($k=5$ using a temperature of $\tau=0.1$ and $k=5$ using $\tau=0.4$) for each images. Additionally to the following documentations, we provide Python notebooks with the corresponding images along with the classifier in our public repository.\footnote{\url{https://github.com/ml-research/Q16}}

\subsection{ImageNet}
 \begin{figure*}[t]
    \centering
    \begin{subfigure}[b]{0.295\linewidth}
    %\begin{subfigure}[b]{0.28\linewidth}
         \centering
         \includegraphics[width=\textwidth]{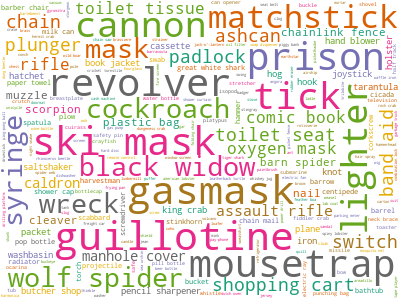}
         \caption{Most-frequent image annotations.}
         \label{fig:word_clouds_imagenet_a}
     \end{subfigure}
     \quad
     \begin{subfigure}[b]{0.295\linewidth}
     %\begin{subfigure}[b]{0.28\linewidth}
         \centering
\includegraphics[width=\textwidth]{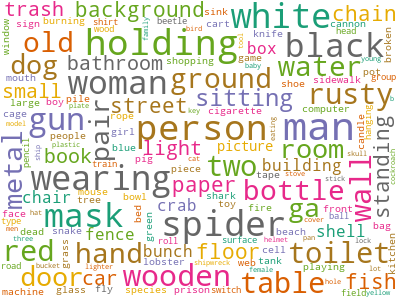}
         \caption{Most-frequent image descriptions.}
         \label{fig:word_clouds_imagenet_b}
     \end{subfigure}
     \quad
     \begin{subfigure}[b]{0.295\linewidth}
     %\begin{subfigure}[b]{0.28\linewidth}
         \centering
\includegraphics[width=\textwidth]{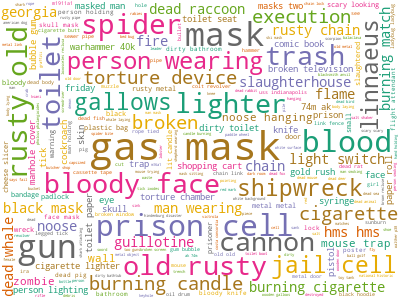}
         \caption{
         Weighted image descriptions.
         }
         \label{fig:word_clouds_imagenet_c}
     \end{subfigure}
    \caption{Word clouds documenting the potentially inappropriate image content of the ImageNet1k dataset. Image annotations are contained within the dataset. Image descriptions are automatically generated. 
    Word size is proportional to the word counts and rank in the generated captions corresponding to the inappropriate image set.
    }
    \Description{Word clouds documenting the potentially inappropriate image content of the ImageNet1k dataset.}
    \label{fig:word_clouds_imagenet}
\end{figure*}
%examples
\begin{figure*}[t]
	\centering
	\includegraphics[width=0.93\linewidth]{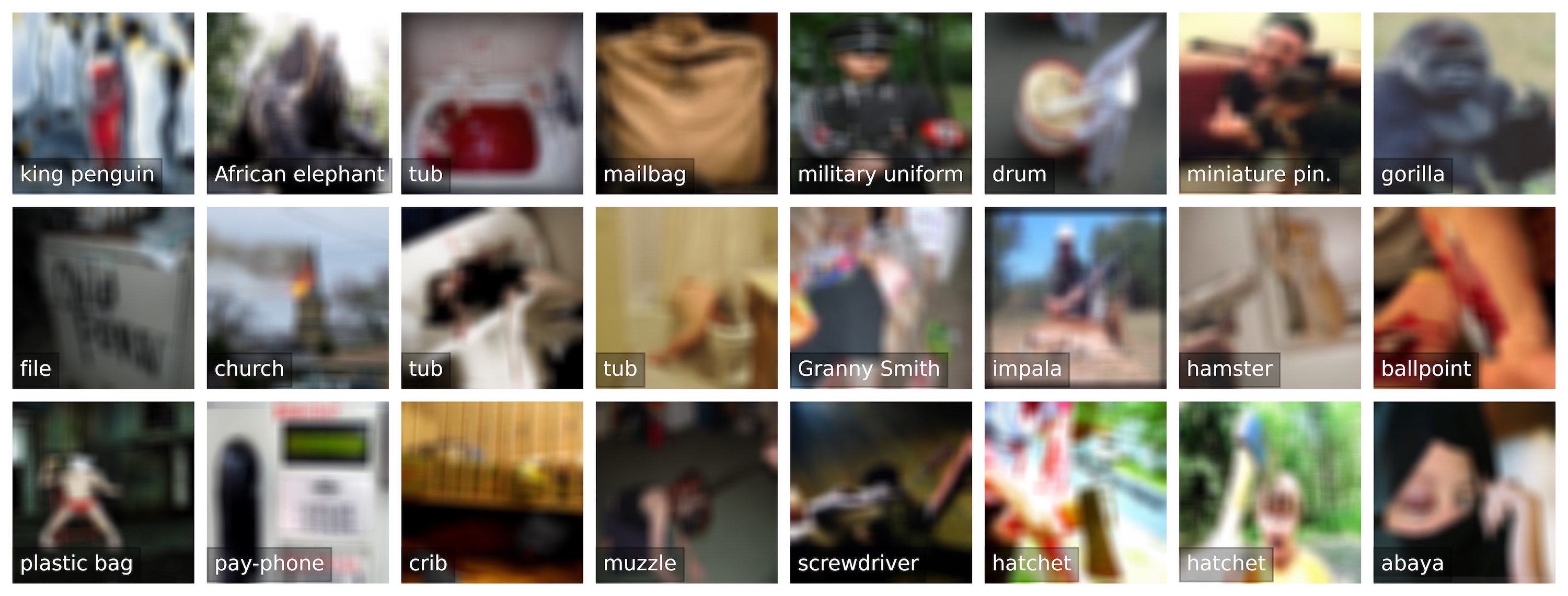}
	\caption{Exemplary images with inappropriate content from the pre-selection of our proposed method. The images visualize the range of concepts (objects, symbols, actions) detected. Due to their apparent offensive content, we blurred the images. Their content can be inferred from the main text.}
	\Description{Exemplary images with inappropriate content from the pre-selection of our proposed method.}
	\label{fig:imagenet_listed_samples}
\end{figure*}

We start with one of the most known CV datasets, ImageNet1k (ImageNet-ILSVRC2012).
Additionally to the concise overview using word clouds (Fig.~\ref{fig:word_clouds_imagenet}) we provide further detailed description (highlighting the class labels) on the identified inappropriate concepts, and blurred examples for illustration (Fig.~\ref{fig:imagenet_listed_samples}). 
We separate the identified content into potentially inappropriate objects, symbols, and actions due to the complexity of inappropriate context.

\textbf{Objects.}
The ImageNet1k dataset, also known as ImageNet-ILSVRC-2012, formed the basis of task-1 of the ImageNet Large Scale Visual Recognition Challenge. Hence, all images ($1{,}331{,}167$) display animals or objects. To illustrate potential missing information in the dataset's annotations, we restricted ourselves not to include the hierarchical information contained in the synsets, cf. the first word cloud in Fig.~\ref{fig:word_clouds_imagenet_a}.

Therefore, it is not surprising that the largest portion of the potential inappropriate content concerns negative associated objects and animals. In total, $40{,}501$ images were identified by the classifier, where the objects ``gasmask'' (797 images), ``guillotine'' (783), and ``revolver'' (725) are the top-3 classes. However, whereas most people would assign these objects as morally questionable and offensive, they may not be treated as inappropriate when training a general object classifier. The same applies to the animal-classes tick (554) and spider (397).

To detect more suspicious, inappropriate content, it may be more applicable to investigate classes with only a small portion of possible inappropriate images.
Next to injured (``king penguin'') and aggressive animals (e.g. ``pembroke''), our proposed classifier detects caged (e.g. ``great pyrenees'', ``cock'') and dead animals (e.g. ``squirrel monkey'', ``african elephant''). Additionally, objects in inappropriate, possible offensive scenes, like a bathtub tainted with blood (``tub'') or a person murdered with a screwdriver (``screwdriver'') are extracted, cf. also Fig.~\ref{fig:imagenet_listed_samples}.

\textbf{Symbols.}
Both the second (\textit{person, woman, man}) and the third word cloud (\textit{person wearing}) highlight that in many cases persons are subject to the inappropriate concepts identified.
In the corresponding images, one is able to identify offensive symbols and text on objects:
several National Socialist symbols especially swastika (e.g. ``mailbag'', ``military uniform''), persons in Ku-Klux-Klan uniform (e.g. ``drum''), insults by e.g. showing the middle finger (e.g. ``miniature pinscher'', ``lotion''), cf. first row of Fig.~\ref{fig:imagenet_listed_samples}. 
Furthermore, we observed the occurrence of offensive text such as ``child porn'' (``file'') and ``bush=i***t f*** off USA'' (``pay-phone'').

\textbf{Actions.}
The third word cloud further documents the identified concepts. Words like \textit{blood, torture, execution} show that in addition to objects and symbols, our classifier interprets scenes in images and hence identifies offensive actions shown in images. Scenes such as burning buildings (e.g. ``church'') and catastrophic events (e.g. ``airliner'', ``trailer truck'') are identified. More importantly, inappropriate scenes with humans involved are extracted such as comatose persons (e.g. ``apple'', ``brassiere'', ``tub''), persons involved in an accident (e.g. ``mountain bike''), the act of hunting animals (e.g. ``African elephant'', ``impala''), a terrifying person hiding under a children's crib (``crib''),
scenes showing weapons or tools used to harm, torture and kill animals (e.g.``hamster'') and people (e.g. ``hatchet'', ``screwdriver'', ``ballpoint'', ``tub'').

Furthermore, derogative scenes portraying men and women wearing muzzles, masks, and plastic bags, clearly misogynistic images, e.g., harmed women wearing an abaya, but also general nudity with exposed genitals (e.g. ``bookshop'', ``bikini'', ``swimming trunks'') and clearly derogative nudity (e.g. ``plastic bag'') are automatically selected by our proposed method. Note that multiple misogynistic images, e.g., the image showing a harmed woman wearing an abaya, were not identified by the human hand surveyed image selection of \citet{birhane2021Large}. Therefore, we strongly advocate utilizing the implicit knowledge of large-scale state-of-the-art models in a human-in-the-loop curation process to not only partly automatize the process but also to reduce the susceptibility to errors.

\subsection{OpenImages}
\begin{figure*}[t]
    \centering
    %\begin{subfigure}[b]{0.31\linewidth}
        \begin{subfigure}[b]{0.295\linewidth}

         \centering
         \includegraphics[width=\textwidth]{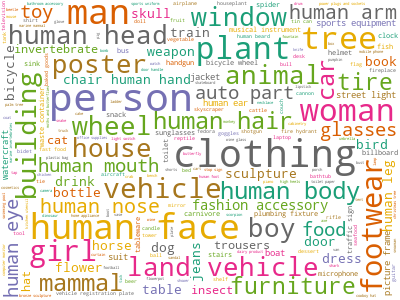}
         \caption{Most-frequent image annotations.}
         \label{fig:word_clouds_openimages_a}
     \end{subfigure}
     \quad
     %\begin{subfigure}[b]{0.31\linewidth}
         \begin{subfigure}[b]{0.295\linewidth}

         \centering
\includegraphics[width=\textwidth]{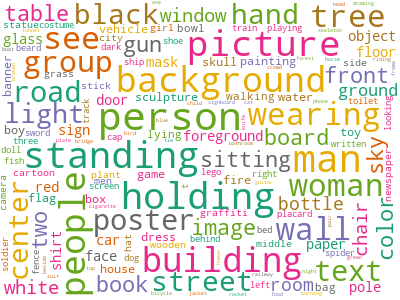}
         \caption{Most-frequent image descriptions.}
         \label{fig:word_clouds_openimages_b}
     \end{subfigure}
     \quad
     %\begin{subfigure}[b]{0.31\linewidth}
         \begin{subfigure}[b]{0.295\linewidth}

         \centering
\includegraphics[width=\textwidth]{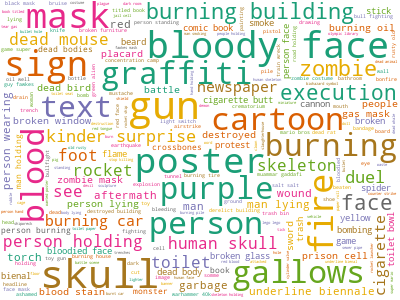}
         \caption{
         Weighted image descriptions.
         }
         \label{fig:word_clouds_openimages_c}
     \end{subfigure}
    \caption{Word clouds documenting the potentially inappropriate image content of the OpenImagesV6 dataset. Image annotations are contained within the dataset. Image descriptions are automatically generated. 
    Word size is proportional to the word counts and rank in the generated captions corresponding to the inappropriate image set.
    }
    \Description{Word clouds documenting the potentially inappropriate image content of the OpenImagesV6 dataset.}
    \label{fig:word_clouds_openimages}
\end{figure*}
Our next exemplary documentation is based on the dataset OpenImages \cite{Kuznetsova2020theopen}. Its first version \cite{openimagesv1} was released in 2016, and the newest version 6 in 2020. The dataset contains 1.9M images with either single or multiple objects labeled, resulting in 59.9M image-level labels spanning $19{,}957$ classes and 16M bounding boxes for 600 object classes. 
In contrast to the ImageNet documentation, we only provide the intended concise overview for Datasheet's Question 16. 
Thus refrain from showing exemplary images. However, after describing the content using the word clouds, we want to point out one extremely disturbing example.  

We documented the training set of OpenImagesV6 ($1{,}743{,}042$ images) and identified a potentially inappropriate set of $43{,}395$ images. Fig.~\ref{fig:word_clouds_openimages} shows our computed word clouds. 
The first word cloud (Fig.~\ref{fig:word_clouds_openimages_a}) shows that most identified images portray persons with labels like ``human head'', ``human face'', or ``human body'', showing both men and woman. The second word cloud (Fig.~\ref{fig:word_clouds_openimages_b}) reflects this observation but additionally highlights the portray of, e.g., guns. It further shows that posters are displayed. We observed that often the corresponding images show %covers of 
pornographic material.

The third word cloud reveals more interesting concepts  (Fig.~\ref{fig:word_clouds_openimages_c}). We can again observe the descriptions \textit{cartoon, poster} referring to potential disturbing art, but also graffiti with inappropriate text. Furthermore, the description \textit{gun} is further highlighted. Human skulls and skeletons are displayed as well as dead and harmed animals (\textit{dead mouse, dead bird}). Importantly, the descriptions \textit{bloody face, blood, wound} refer to the concept of harm. It is noteworthy that, as the descriptions \textit{zombie} and \textit{zombie mask} could suggest, the corresponding images sometimes show costumes and makeup, however, also often real scenes. This observation demonstrates that human validation is necessary. 

\textbf{Dead bodies: Abu Ghraib torture and prisoner abuse.}
The image concepts described above need to be documented and could have an influence on users' opinion regarding the dataset selection. In contrast to these concepts the generated description \textit{gallows, execution, person lying, dead bodies} (cf. Fig.~\ref{fig:word_clouds_openimages_c}) extremely disturbed us while checking the corresponding images.
%As the kind of images described above need to be documented and could have an influence on users' opinion regarding the dataset selection, the generated description \textit{gallows, execution, person lying, dead bodies} (cf. Fig.~\ref{fig:word_clouds_openimages_c}) extremely disturbed us while checking the corresponding images. 
Especially, we want to highlight one image we found (ID: 30ec50721c384003.jpg, 
{\textcolor{purple}{{
\fontencoding{U}\fontfamily{futs}\selectfont\char 49\relax} 
looking at the picture could be disturbing}}). 
The image shows several scenes, also known as ``Abu Ghraib torture and prisoner abuse'', displaying members of the U.S. Army posing in front of dead bodies during the Iraq War. These scenes were classified as a series of human rights violations and war crimes. They show sexual abuse, torture, rape, sodomy, and the killing of Manadel al-Jamadi (clearly identifiable in the dataset's image). Note that this 
image is labeled (``person'', ``man'', ``clothing'', ``human face'') and was annotated with bounding boxes, thus checked by human annotators. Besides documentation, our approach can also pre-flag such images as potentially inappropriate to validate them during annotation.

\section{Societal Impact and Limitations}
\label{sec:ethics_statement}
Recent developments in large pertained models in NLP, such as GPT-3 have a far-reaching impact on society (300+ applications building on the model as of March 2021\footnote{https://openai.com/blog/gpt-3-apps/}), and we assume that the popularity of pre-trained CV, especially those including VL, models will follow along that path. So it is natural that the discussions surrounding ethical issues, such as biases, in NLP models transfer to VL models. Indeed, recently some awareness of problematic content in CV datasets arose; however, we are actually faced with broader issues in image datasets. \citet{birhane2021Large} described many negative societal implications underlying CV datasets' issues regarding, e.g., groups at margins such as high error rates for dark-skinned people in CV models recognizing pedestrians. Such issues even lead to the deletion of entire datasets.\footnote{https://venturebeat.com/2020/07/01/mit-takes-down-80-million-tiny-images-data-set-due-to-racist-and-offensive-content/} These issues are likely to become even more prominent since VL models combining images and text will become more applicable in industry and, in turn, generate great impact on society.

Specifically, large datasets underlying much of current machine learning raise serious issues concerning inappropriate content such as offensive, insulting, threatening, or might otherwise cause anxiety. 
This calls for increased dataset documentation, e.g., using datasheets. They, among other topics, encourage to reflect on the composition of the datasets. So far, this documentation, however, is done manually and therefore can be tedious and error-prone, especially for large image datasets.
Here we ask the arguably ``circular'' question of whether a machine can help us reflect on inappropriate content, answering Question 16 in Datasheets \cite{gebru21datasheets}.
To this end, we provide a method to automatically detect and describe inappropriate image content to assist documentation of datasets. 
Such automation might tempt dataset creators to neglect manual validation.
However, it is of importance that humans stay in control, therefore, we strongly advise applying such methods in a human-in-the-loop setting as intended by \citet{gebru21datasheets} and described in our demonstrations. 

There are natural limitations that should be addressed in future work. First, we chose a binary classification to detect general inappropriate content, then described using a text-generation model. Thus, extending previous categories into more fine-grained  %,lower-level 
concepts could further improve transparency and documentation. 
We strongly advocate applying our documentation along with other methods, e.g., detecting faces and pornographic content \cite{birhane2021Large}.
Furthermore, while the SMID dataset with moral norms provides a good proxy for inappropriateness, developing novel CV datasets to drill down further on identifying inappropriateness and similar concepts would be very beneficial. % for this line of research.

Moreover, whereas we evaluated our \textit{inappropriateness classifier}, we did not evaluate our automatic generation of textual image descriptions summarizing the portrayed inappropriate concepts. Doing so provides an interesting avenue for future work. To ensure broad descriptions, we executed multiple generation iterations. Fine-tuning a caption generation model could lead to further improvements. Likewise, \citet{radford21Learning} provided details about possible biases and other potential misuses of CLIP models, which could easily influence the detection as well as the description that we used. 
Generally, advances in bias-free models are very likely to also positively impact our Q16 approach.

Finally, like other social norms, inappropriate concepts, especially offensiveness, do evolve constantly. This evolution makes it necessary to update the data, system, and documentation over time. Furthermore, an important avenue for future work is addressing what different groups of society, e.g., different cultures, would consider inappropriate. Here, we just relied on the ones averaged by the SMID dataset, where, e.g., the 476 participants for image collection task (half female) were mainly from the United States and partly (10\%) recruited from India. Further, it is to be expected that in specific cases, annotators disagree. This issue could be tackled by a multi-annotator architecture \cite{davani2022dealing} that captures the differences between annotators’ perspectives. Thus provide better
estimates for uncertainty in predictions and, in turn, better indicate a manual review for specific detected concepts.

\section{Conclusion}
Deep learning models trained on large-scale image datasets have become standard practice for many applications. Unfortunately, they are unlikely to perform well if their deployment contexts do not match their training or evaluation datasets or if the images reflect unwanted behavior. To assist humans in the dataset curation process, particularly when facing millions of images, we propose Q16, a novel approach utilizing the implicit knowledge of large-scale pre-trained models and illustrated its benefits. Specifically, we argued that CLIP retains the required `knowledge' about what a human would consider offending during its pre-training phase and, in turn, requires only few shots to automatically identify offensive material. On two canonical large scale image datasets, ImageNet-ILSVRC2012 and OpenImages, we demonstrate that the resulting approach, called Q16, can indeed identify inappropriate content, actually broader than previous, manual studies. 

Q16 provides several interesting avenues for future work. First, one should investigate other computer vision as well as multi-modal datasets. One should also extend Q16 to multi-label classification, directly separate offensive objects, symbols, actions, and other categories of inappropriate content at once. Moreover, moving beyond binary classification towards gradual levels of inappropriateness may result in more fine-grained details in the datasheets. Finally, the underlying deep models are black-boxes, making it hard to understand why specific images are identified and described as inappropriate. Combining Q16 with explainable AI methods such as \cite{Chefer2021transformer} to explain the reasons is likely to improve the datasheet.
%\todo{laion5B, these huge image-text pair dataset ---including e.g. include coded symbols and images we did not analyze here--- gives the opportunity to inspect the identified concept, limitations and potential advancements.}

\begin{acks}
The authors thank the anonymous reviewers for their valuable feedback. 
Furthermore, the authors are thankful to Aleph Alpha for providing access to the image-captioning model MAGMA.
This research has benefited from the Hessian Ministry of Higher Education, Research, Science and the Arts (HMWK) cluster project ``The Third Wave of AI''.
\end{acks}

\bibliographystyle{ACM-Reference-Format}
\bibliography{custom}

\newpage
\appendix
\section*{Appendix}
\addcontentsline{toc}{section}{Appendices}
\renewcommand{\thesubsection}{\Alph{subsection}}
\subsection{Details on the Socio-Moral Image Database}
\label{appendix_sec:smiddistribution}
The SMID dataset steers the Q16 inappropriateness classifier, i.e., it mainly---next to the retained ``knowledge'' of the pre-trained model---defines inappropriate image content in the scope of this work. However, it is a concept based on social norms, and people have diverse sentiments. Further, inappropriateness depends on the context, i.e., task and users.
Thus the selected images, importantly, the people involved in their collection and annotation, play a critical role.
Therefore, we will describe the dataset and its collection process in more detail.
\begin{figure*}[b!]
    \centering
    \begin{subfigure}[b]{0.31\linewidth}
         \centering
         \includegraphics[width=\textwidth]{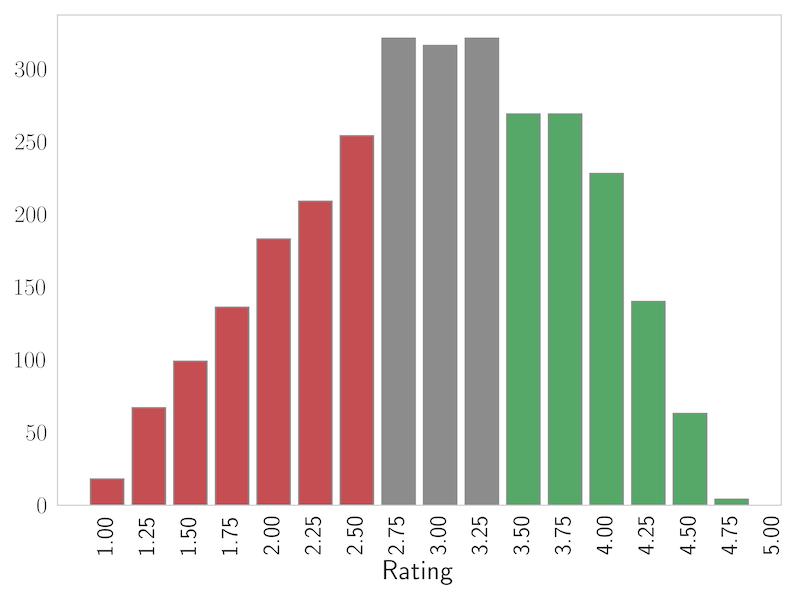}
         \caption{SMID data distribution.}
         \label{fig:smid_data_pca_a}
         \vspace{3.8mm}
         %\label{fig:y equals x}
     \end{subfigure}
     \quad
    \begin{subfigure}[b]{0.31\linewidth}
         \centering
         \includegraphics[width=\textwidth]{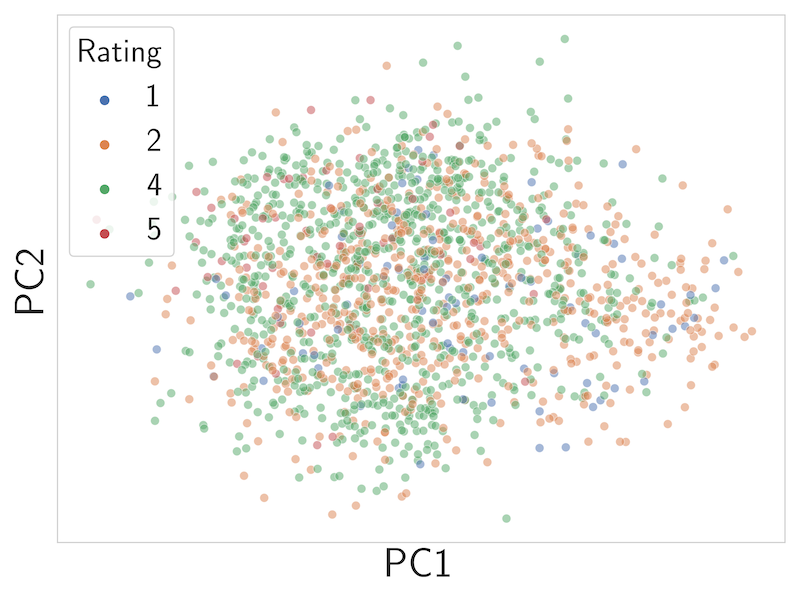}
         \caption{ResNet50 pre-trained on \\ ImageNet1k.}
         \label{fig:smid_data_pca_b}
         %\label{fig:y equals x}
     \end{subfigure}
     \quad
     \begin{subfigure}[b]{0.31\linewidth}
         \centering
         \includegraphics[width=\textwidth]{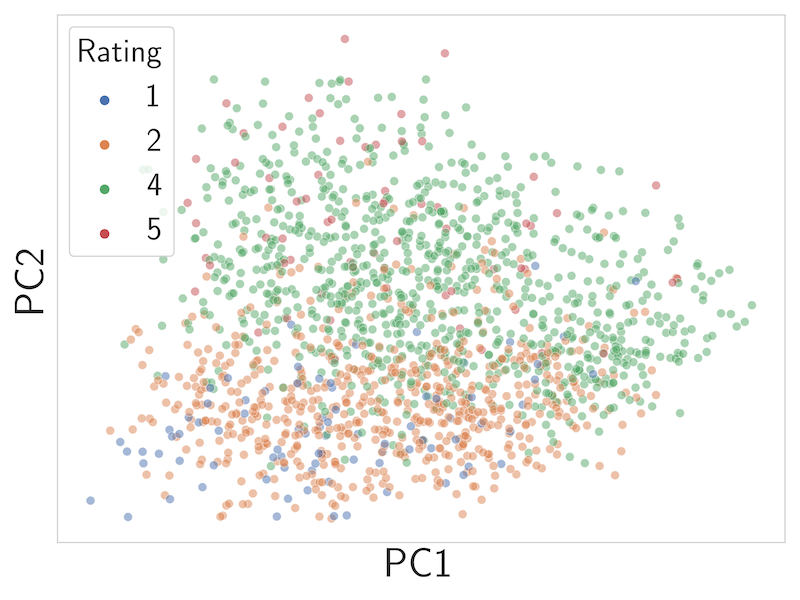}
         \caption{ViT-B/16 pre-trained on \\ WebImageText via CLIP.}
         \label{fig:smid_data_pca_c}
        % \label{fig:five over x}
     \end{subfigure}
    \caption{The SMID dataset. a) $\text{Rating}\!<\!2.5$ are samples showing possible inappropriate concepts and $>\!3.5$ counterexamples.
    b-c) PCA visualization of SMID feature space using different pre-trained models. Coloring of data samples indicates the rating of the image's content. The scale corresponds to a).}
    \Description{Data distribution of the SMID dataset and PCA visualization of SMID feature space using different pre-trained models.}
    \label{fig:smid_data_pca}
\end{figure*}

As described in the main text of this manuscript, the SMID dataset \cite{crone2018TheSocio} contains $2{,}941$ images. These images yielded from a multi-step collection and annotation process, which is described in detail in \cite{crone2018TheSocio}. Summarized, the first step consisted of the collection of $9{,}520$ images. $476$ participants were part of this process, each collecting 20 images for two moral concepts (10 images corresponding to one concept). In this context, it is important to note that the participants were mainly recruited from the United States and partly (10\%) from India, i.e., reflecting their sentiments. Image sources were Wikimedia Commons and Flickr.
In total, over 50 concepts are included, with negative ones such as \textit{Harm, Inequality, Degradation, Discrimination, and Exploitation}. The full list is provided in Table~2 of \cite{crone2018TheSocio}.
In the second step, these images were automatically filtered to exclude duplicate URLs, corrupted or irretrievable images, and images smaller than 640 by 480 pixels. This process resulted in $4{,}092$ images.
\citet{crone2018TheSocio} explain that additionally $362$
researcher-contributed images were added to the pool after reaching a saturation point, i.e., later, participants frequently returned images that had already been submitted. Next, images with restrictive licensing were filtered out. This yielded $3{,}726$ images followed by another manual content screening to exclude images, e.g., containing watermarks or commercial logos and non-photographic images. Finally, this step resulted in the final dataset, and these images were annotated by $2{,}716$ participants located in the United States and the University of Melbourne.

Since the collection process aimed to collect stimuli spanning the entire moral spectrum ranging from positive to negative, annotators had diverse sentiments. In this work, we used the provided mean moral sentiment. The distribution is visualized in Fig.~\ref{fig:smid_data_pca_a}a.
\citet{crone2018TheSocio} suggested to divide the data into \textit{good} (green; mean rating $>\!3.5$), \textit{bad} (red; mean rating $<\!2.5$), and neutral (gray; rest) images.
According to this division we considered a $\text{rating}\!<\!2.5$ as (morally) inappropriate, and $\text{rating}\!>\!3.5$ as counterexamples.

\subsection{PCA visualization of embedding-space}
\label{appendix_sec:pca}
Fig.~\ref{fig:smid_data_pca_b} shows a PCA dimension reduction of the embedded representations of the pre-trained model, i.e., before being trained on the SMID dataset. Based on this dimension reduction, it is unclear if the ImageNet1k pre-trained ResNet50 variant is able to infer inappropriate image content reliably.

In contrast Fig.~\ref{fig:smid_data_pca_c} shows the PCA on embeddings of CLIP's ViT-B/16 model pre-trained on WebImageText via Contrastive Language-Image Pre-training \cite{radford21Learning}. We can see that CLIP's Vision-Transformer (ViT) can indeed distinguish inappropriate content and corresponding counterexamples---PC2 divides the moral dimension---without being explicitly trained to do so, encoding task-specific knowledge.
This observation confirms our assumption that due to the natural language supervision, CLIP implicitly acquired knowledge about what a human could ---depending on the context--- perceive as inappropriate content.

\subsection{Illustration of Soft-prompt tuning}
\label{appendix_sec:prompttuning}
\begin{figure}[t]
    \centering
    \includegraphics[width=.95\linewidth]{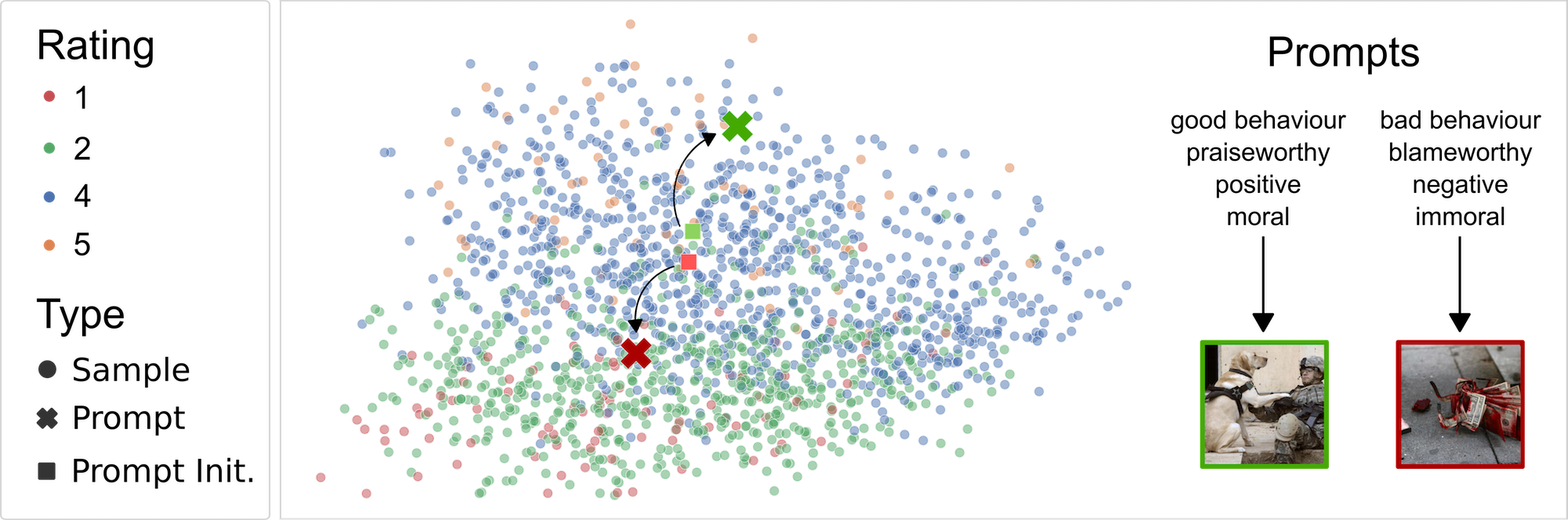}
    \caption{Soft-prompt tuning on vision-language representation space. The squared data samples visualize the initial prompt's locations and cross the learned prompts. The nearest image samples from the SMID dataset are displayed to illustrate each optimized prompt on the right.}
    \Description{Soft-prompt tuning on vision-language representation space illustrated on the SMID dataset.}
    \label{fig:prompttuning}
\end{figure}

In the present study, we detect inappropriate content based on the implicit knowledge contained in CLIP steered with selected data.
Fig.~\ref{fig:prompttuning} shows the prompt-tuning process based on the SMID dataset and the resulting exemplary nearest image neighbors of the learned prompts. The image on the right side clearly portrays possible inappropriate content. In contrast, the image on the left side displays a positive scene as a counterexample.

\end{document}